\documentclass[11pt]{article}
\usepackage{eacl2017}
\usepackage{times}
\usepackage{url}
\usepackage{latexsym}
\usepackage{amsmath, amsfonts, amssymb}
\usepackage[svgnames]{xcolor}
\usepackage{comment}
\usepackage{caption}
\usepackage{xspace}
\usepackage{lipsum}
\usepackage{multirow}

\eaclfinalcopy

\newcommand{\modelname}{neural readers\xspace}
\newcommand{\train}{\textsc{train}\xspace}
\newcommand{\val}{\textsc{val}\xspace}
\newcommand{\control}{\textsc{control}\xspace}
\newcommand{\dev}{\textsc{dev}\xspace}
\newcommand{\test}{\textsc{test}\xspace}
\newcommand{\ngram}{$n$-gram\xspace}

\newenvironment{itemizesquish}{\begin{list}{\labelitemi}{\setlength{\itemsep}{0em}\setlength{\labelwidth}{0.5em}\setlength{\leftmargin}{\labelwidth}\addtolength{\leftmargin}{\labelsep}}}{\end{list}}

\newenvironment{enumeratesquish}{\begin{list}{\addtocounter{enumi}{1}\labelenumi}{\setlength{\itemsep}{0em}\setlength{\labelwidth}{0.5em}\setlength{\leftmargin}{\labelwidth}\addtolength{\leftmargin}{\labelsep}}}{\end{list}\setcounter{enumi}{0}}

\DeclareMathOperator*{\argmax}{argmax}

\title{Broad Context Language Modeling as Reading Comprehension}

\author{Zewei Chu$^1$ \ \ \ \ \ \  Hai Wang$^2$ \ \ \ \ \ \ Kevin Gimpel$^2$ \ \ \ \ \ \ David McAllester$^2$ \\
[1ex]
$^1$University of Chicago, Chicago, IL, 60637, USA\\
$^2$Toyota Technological Institute at Chicago, Chicago, IL, 60637, USA\\
[1ex]
{\small
{\tt zeweichu@uchicago.edu},  
{\tt \{haiwang,kgimpel,mcallester\}@ttic.edu}
}\\}

\date{}

\begin{document}
\maketitle
\begin{abstract}
Progress in text understanding has been driven by large datasets that test particular capabilities, like recent datasets for reading comprehension~\cite{deepmind-reader:15}. We focus here on the LAMBADA dataset~\cite{lambada:16}, a word prediction task requiring broader context than the immediate sentence. We view LAMBADA as a reading comprehension problem and apply comprehension models based on neural networks. Though these models are constrained to choose a word from the context, they improve the state of the art on LAMBADA from 7.3\% to 49\%. We analyze  100 instances, finding that neural network readers perform well in cases that involve selecting a name from the context based on dialogue or discourse cues but struggle when coreference resolution or external knowledge is needed. 
\end{abstract}

\section{Introduction}

The LAMBADA dataset~\cite{lambada:16} was designed by identifying word prediction tasks that require broad context.  Each  instance is drawn from the BookCorpus~\cite{zhu:15} and consists of a passage of several sentences where the task is to predict the last word of the last sentence.  
The instances are manually filtered to find cases that are guessable by humans when given the larger context but not when only given the last sentence.  The expense of this manual filtering has limited the dataset to only about 10,000 instances which are viewed as development and test data.  The training data is taken to be books in the corpus other than those from which the evaluation passages were extracted. 

\newcite{lambada:16} provide baseline results with popular language models and neural network architectures; all achieve zero percent accuracy. 
The best accuracy is 7.3\% obtained by randomly choosing a capitalized word from the passage.

Our approach is based on the observation that in 83\% of instances the answer appears in the context. We exploit this in two ways.  First, we automatically construct a large training set of 1.8 million instances by simply selecting passages where the answer occurs in the context.  Second, we treat the problem as a reading comprehension task similar to the CNN/Daily Mail datasets introduced by \newcite{deepmind-reader:15}, the Children's Book Test (CBT) of \newcite{cbt:15}, and the Who-did-What dataset of \newcite{onishi-16-full}.  We show that standard models for reading comprehension, trained on our automatically generated training set, improve the state of the art on the LAMBADA test set from 7.3\% to 49.0\%.  This is in spite of the fact that these models fail on the 17\% of instances in which the answer is not in the context. 

We also perform a manual analysis of the LAMBADA task, provide an estimate of human performance, and categorize the instances in terms of the phenomena they test. We find that the comprehension models perform best on instances that require selecting a name from the context based on dialogue or discourse cues, but struggle when required to do coreference resolution or when  external knowledge could help in choosing the answer. 

\section{Methods}

We now describe the models that we employ for the LAMBADA task (Section~\ref{sec:readers}) as well as our dataset construction procedure (Section~\ref{sec:data}). 

\subsection{Neural Readers}
\label{sec:readers}
\newcite{deepmind-reader:15} developed the CNN/Daily Mail comprehension tasks and introduced question answering models based on neural networks. Many others have been developed since. 
We refer to these models as ``\modelname''. While a detailed survey is beyond our scope, we briefly describe the \modelname used in our experiments: the Stanford~\cite{danqi-reader:16}, Attention Sum~\cite{as-reader:16}, and Gated-Attention~\cite{gated-as-reader:16} Readers. These \modelname use attention based on the question and passage to choose an answer from among the words in the passage. 
We use $\textbf{d}$ for the context word sequence, $\textbf{q}$ for the question  (with a blank to be filled), $\mathcal{A}$ for the candidate answer list, and $\mathcal{V}$ for the vocabulary. We describe neural readers in terms of three components:

\begin{enumeratesquish}

\item \textbf{Embedding and Encoding}\label{embedding}: Each word in $\textbf{d}$ and $\textbf{q}$ is mapped into a $v$-dimensional vector via the embedding function $e(w) \in \mathbb{R}^v$, for all $w \in \textbf{d} \cup \textbf{q}$.\footnote{We overload the $e$ function to operate on sequences and denote the embedding of $\textbf{d}$ and $\textbf{q}$ as matrices $e(\textbf{d})$ and $e(\textbf{q})$.} 
The same embedding function is used for both $\textbf{d}$ and $\textbf{q}$. The embeddings are learned from random initialization; no pretrained word embeddings are used. 
The embedded context is processed by a bidirectional recurrent neural network (RNN) which computes hidden vectors $h_i$ for each position $i$: 
\begin{align}
h^{\rightarrow} &= \mathit{fRNN}(\boldsymbol{\theta}_d^\rightarrow, e(\textbf{d}))\nonumber \\
h^{\leftarrow} &= \mathit{bRNN}(\boldsymbol{\theta}_d^\leftarrow, e(\textbf{d}))\nonumber \\
h &= \langle h^{\rightarrow}, h^{\leftarrow}\rangle \nonumber
\end{align}
\noindent where $\boldsymbol{\theta}_d^\rightarrow$ and $\boldsymbol{\theta}_d^\leftarrow$ are RNN parameters, and each of $\mathit{fRNN}$ and $\mathit{bRNN}$ return a sequence of hidden vectors, one for each position in the input $e(\boldsymbol{d})$. 
The question is encoded into a single vector $g$ which is the concatenation of the final vectors of two RNNs:
\begin{align}
g^{\rightarrow} &= \mathit{fRNN}(\boldsymbol{\theta}_q^\rightarrow, e(\textbf{q}))\nonumber \\
g^{\leftarrow} &= \mathit{bRNN}(\boldsymbol{\theta}_q^\leftarrow, e(\textbf{q}))\nonumber \\
g &= \langle g_{|\boldsymbol{q}|}^{\rightarrow}, g_0^{\leftarrow} \rangle\nonumber
\end{align}
\noindent The RNNs use either gated recurrent units~\cite{cho-gru:14} or long short-term memory~\cite{lstm:97}. 

\item \textbf{Attention}: The readers then compute attention weights on positions of $h$ using $g$. 
In general, we define 
$\alpha_{i} = \mathrm{softmax}(\mathit{att}(h_i, g))$, 
where $i$ ranges over positions in $h$. 
The $\mathit{att}$ function is an inner product in the Attention Sum Reader and a bilinear product in the Stanford Reader. 
The computed attentions are then passed through a $\mathrm{softmax}$ function to form a probability distribution. 
The Gated-Attention Reader uses a richer attention architecture~\cite{gated-as-reader:16}; space does not permit a detailed description. 

\item \textbf{Output and Prediction}: To output a prediction $a^\ast$, 
the Stanford Reader computes the attention-weighted sum of the context vectors and then an inner product with each candidate answer: 
\begin{align}
\textbf{c} = \sum_{i=1}^{|\textbf{d}|} \alpha_i h_i\quad\;\; 
a^\ast = \argmax_{a \in \mathcal{A}}\; o(a)^\top 
\textbf{c}\nonumber
\end{align}
\noindent where $o(a)$ is the ``output'' embedding function. 
As the Stanford Reader was developed for the anonymized CNN/Daily Mail tasks, only a few entries in the output embedding function needed to be well-trained in their experiments. However, for LAMBADA, correct answers can range over the entirety of $\mathcal{V}$, making the output embedding function difficult to train. Therefore we also experiment with a modified version of the Stanford Reader that uses the same embedding function $e$ for both input and output words:
\begin{equation}
a^\ast = \argmax_{a \in \mathcal{A}} \;e(a)^{\top} W \textbf{c}\label{eq:mod}
\end{equation}
where $W$ is an additional parameter matrix used to match  dimensions and model any additional needed transformation.  

For the Attention Sum and Gated-Attention Readers the answer is computed by: 
\begin{align}
\forall a\in \mathcal{A}, \:P(a|\textbf{d}, \textbf{q}) &= \sum_{i \in I(a, \textbf{d})} \alpha_i \nonumber\\
a^* &= \argmax_{a \in \mathcal{A}} \;P(a|\textbf{d}, \textbf{q})\nonumber
\end{align}
where $I(a, \textbf{d})$ is the set of positions where $a$ appears in context $\textbf{d}$. 
\end{enumeratesquish}

\subsection{Training Data Construction}
\label{sec:data}

Each LAMBADA instance is divided into a \textbf{context} (4.6 sentences on average) and a \textbf{target sentence}, and the last word of the target sentence is the \textbf{target word} to be predicted. The LAMBADA dataset consists of development (\dev) and test (\test) sets; \newcite{lambada:16} also provide 
a control dataset (\control), an unfiltered sample of instances from the BookCorpus. 

We construct a new training dataset from the BookCorpus. 
We restrict it to instances that contain the target word in the context. 
This decision is natural given our use of neural readers that assume the answer is contained in the passage. We also ensure that the context has at least 50 words and contains 4 or 5 sentences and we require the target sentences to have more than 10 words. 

Some neural readers require a candidate target word list to choose from. 
We list all words in the context as candidate answers, except for punctuation.\footnote{This list of punctuation symbols is at \url{https://raw.githubusercontent.com/ZeweiChu/lambada-dataset/master/stopwords/shortlist-stopwords.txt}}
Our new dataset contains 1,827,123 instances in total. We divide it into two parts, a training set (\train) of 1,618,782 instances and a validation set (\val) of 208,341 instances. These datasets can be found at the authors' websites. 

\section{Experiments}

We use the Stanford Reader~\cite{danqi-reader:16}, our modified Stanford Reader (Eq.~\ref{eq:mod}), the Attention Sum (AS) Reader~\cite{as-reader:16}, and the Gated-Attention (GA) Reader~\cite{gated-as-reader:16}. 
We also add the simple features from \newcite{hai-logical:16} to the AS and GA Readers. 
The features are concatenated to the word embeddings in the context. They include: whether the word appears in the target sentence, the frequency of the word in the context, the position of the word's first occurrence in the context as a percentage of the context length, and whether the text surrounding the word matches the text surrounding the blank in the target sentence. For the last feature, we only consider matching the left word since the blank is always the last word in the target sentence.

All models are trained end to end without any warm start and without using pretrained embeddings. 
We train each reader on \train for a max of 10 epochs, stopping when accuracy on \dev decreases two epochs in a row. We take the model from the epoch with max \dev accuracy and  evaluate it on \test and \control. \val is not used.

We evaluate several other baseline systems inspired by those of \newcite{lambada:16}, but we focus on versions that restrict the choice of answers to non-stopwords in the context.\footnote{We use the stopword list from \newcite{richardson2013mctest}.}
We found this strategy to consistently improve performance even though it limits the maximum achievable accuracy. 

We consider two \ngram language model baselines. We use the SRILM toolkit~\cite{stolcke-02-srilm} to estimate a 4-gram model with modified Kneser-Ney smoothing on the combination of \train and \val. 
One uses a cache size of 100 and the other does not use a cache. We use each model to score each non-stopword from the context. 
We also evaluate an LSTM language model. 
We train it on \train, where the loss is cross entropy summed over all positions in each instance. The output vocabulary is the vocabulary of \train, approximately 130k word types. At test time, we again limit the search to non-stopwords in the context. 

We also test simple baselines that choose particular non-stopwords from the context, including a random one, the first in the context, the last in the context, and the most frequent in the context. 

\section{Results}

\begin{table}[t]
\small
\begin{center}
\begin{tabular}{|l|r|rr|}
\hline 
\multirow{2}{*}{Method} & \multicolumn{1}{|c|}{\test} & \multicolumn{2}{|c|}{\control} \\
 & \multicolumn{1}{c}{all} & \multicolumn{1}{|c}{all} & \multicolumn{1}{c|}{context} \\
\hline
\multicolumn{4}{|l|}{Baselines \cite{lambada:16}}\\
\hline
Random in context & 1.6 & 0 &  N/A \\ 
Random cap.~in context & 7.3 & 0 &  N/A \\
\ngram & 0.1 & 19.1 & N/A  \\
\ngram + cache & 0.1 & 19.1 & N/A \\
LSTM & 0 & \textbf{21.9} & N/A \\
Memory network & 0 & 8.5 & N/A \\
\hline
\multicolumn{4}{|l|}{Our context-restricted non-stopword baselines} \\
\hline
Random & 5.6 & 0.3 & 2.2 \\
First & 3.8 & 0.1 & 1.1 \\
Last & 6.2 & 0.9 & 6.5 \\
Most frequent & 11.7 & 0.4 & 8.1 \\
\hline
\multicolumn{4}{|l|}{Our context-restricted language model baselines} \\
\hline
\ngram & 10.7 & 2.2 & 15.6 \\
\ngram + cache & 11.8 & 2.2 & 15.6 \\
LSTM &9.2 & 2.4 & 16.9 \\
\hline
\multicolumn{4}{|l|}{Our neural reader results} \\
\hline
Stanford Reader &  21.7 & 7.0 & 49.3 \\
Modified Stanford Reader & 32.1 & 7.4 & 52.3 \\
AS Reader & 41.4 & 8.5 & 60.2 \\
AS Reader + features & 44.5 & 8.6 & 60.6 \\
GA Reader & 45.4 & 8.8 & 62.5 \\
GA Reader + features & \bf 49.0 & 9.3 & \bf 65.6 \\
\hline
Human & {\it 86.0}$^\ast$ & {\it 36.0}$^\dagger$ & - \\
\hline
\end{tabular}
\end{center}
\caption{\label{accuracy-table} Accuracies on \test and \control datasets, computed over all instances (``all'') and separately on those in which the answer is in the context (``context''). The first section is from \newcite{lambada:16}. $^\ast$Estimated from 100 randomly-sampled \dev instances. $^\dagger$Estimated from 100 randomly-sampled \control instances. 
} 
\vspace{-0.4cm}
\end{table}

Table~\ref{accuracy-table} shows our results. 
We report accuracies on the entirety of \test and \control (``all''), as well as separately on the part of \control where the target word is in the context (``context''). The first part of the table shows results from \newcite{lambada:16}. 
We then show our baselines that choose a word from the context. 
Choosing the most frequent yields a surprisingly high accuracy of 11.7\%, which is better than all results from Paperno et al.

Our language models perform comparably, with the $n$-gram + cache model doing best. By forcing language models to select a word from the context, the accuracy on \test is much higher than the analogous models from Paperno et al., though accuracy suffers on \control. 

We then show results with the neural readers, showing that they give much higher accuracies on \test than all other methods. The GA Reader with the simple additional features~\cite{hai-logical:16} yields the highest accuracy, reaching 49.0\%. 
We also measured the ``top $k$'' accuracy of this model, where we give the model credit if the correct answer is among the top $k$ ranked answers. On \test, we reach 65.4\% top-2 accuracy and 72.8\% top-3.

The AS and GA Readers work much better than the Stanford Reader. One cause appears to be that the Stanford Reader learns distinct embeddings for input and answer words, as discussed above. 
Our modified Stanford Reader, which uses only a single set of word embeddings, improves by 10.4\% absolute. 
Since the AS and GA Readers merely score words in the context, they do not learn separate answer word embeddings and therefore do not suffer from this effect. 

We suspect the remaining accuracy difference between the Stanford and the other readers is due to the difference in the output function. The Stanford Reader was developed for the CNN and Daily Mail datasets, in which correct answers are anonymized entity identifiers which are reused across  instances. Since the identifier embeddings are observed so frequently in the training data, they are frequently updated. In our setting, however, answers are words from a large vocabulary, so many of the word embeddings of correct answers may be undertrained. 
This could potentially be addressed by augmenting the word embeddings with identifiers to obtain some of the modeling benefits of anonymization~\cite{hai-logical:16}.

All context restricted models yield poor accuracies on the entirety of \control. This is due to the fact that only 14.1\% of \control instances have the target word in the context, so this sets the upper bound that these models can achieve.

\subsection{Manual Analysis}

One annotator, a native English speaker, sampled 100 instances randomly from \dev, hid the final word, and attempted to guess it from the context and target sentence. The annotator was correct in 86 cases. For the subset that contained the answer in the context, the annotator was correct in 79 of 87 cases. Even though two annotators were able to correctly answer all LAMBADA instances during dataset construction~\cite{lambada:16}, our results give an estimate of how often a third would agree. 
The annotator did the same on 100 instances randomly sampled from \control, guessing correctly in 36 cases. These results are reported in Table~\ref{accuracy-table}. The annotator was correct on 6 of the 12 \control instances in which the answer was contained in the context. 

\begin{table}[t]
\begin{center}
\begin{tabular}{|lr|r|r|}
\hline 
label & \# & \multicolumn{1}{|c|}{GA+} & \multicolumn{1}{|c|}{human} \\
\hline
single name cue & 9 & 89\% & 100\% \\
simple speaker tracking & 19 & 84\% & 100\%\\
basic reference & 18 & 56\% & 72\% \\
discourse inference rule & 16 & 50\% & 88\% \\
semantic trigger & 20 & 40\% & 80\% \\
coreference & 21 & 38\% & 90\% \\
external knowledge & 24 & 21\% & 88\% \\
\hline
all & 100 & 55\% & 86\% \\
\hline
\end{tabular}
\end{center}
\caption{\label{tab:categories} Labels derived from manual analysis of 100 LAMBADA \dev instances. An instance can be tagged with multiple labels, hence the sum of instances across labels exceeds 100.}
\end{table}

We analyzed the 100 LAMBADA \dev instances, tagging each with labels indicating the minimal kinds of understanding needed to answer it correctly.\footnote{The annotations are available from the authors' websites.} 
Each instance can have multiple labels. We briefly describe each label below:
\begin{itemizesquish}
\item single name cue: the answer is clearly a name according to contextual cues and only a single name is mentioned in the context.
\item simple speaker tracking: instance can be answered merely by tracking who is speaking without understanding what they are saying. 
\item basic reference: answer is a reference to something mentioned in the context; simple understanding/context matching suffices.
\item discourse inference rule: answer can be found by applying a single discourse inference rule, such as the rule: 
``$X$ left $Y$ and went in search of $Z$'' $\rightarrow Y \neq Z$. 
\item semantic trigger: amorphous semantic information is needed to choose the answer, typically related to event sequences or dialogue turns, e.g., a customer says ``Where is the $X$?'' and a supplier responds ``We got plenty of $X$''.
\item coreference: instance requires non-trivial coreference resolution to solve correctly, typically the resolution of anaphoric pronouns. 
\item external knowledge: some particular external knowledge is needed to choose the answer.
\end{itemizesquish}
Table~\ref{tab:categories} shows the breakdown of these labels across instances, as well as the accuracy on each label of the GA Reader with features. 

The GA Reader performs well on instances involving shallower, more surface-level cues. 
In 9 cases, the answer is clearly a name based on contextual cues in the target sentence and there is only one name in the context; the reader answers all but one correctly. 
When only simple speaker tracking is needed (19 cases), the reader gets 84\% correct. 

The hardest instances are those that involve deeper understanding, like semantic links, coreference resolution, and external knowledge. While external knowledge is difficult to define, we chose this label when we were able to explicitly write down the knowledge that one would use when answering the instances, e.g., one instance requires knowing that ``when something explodes, noise emanates from it''. These instances make up nearly a quarter of those we analyzed, making LAMBADA a good task for work in leveraging external knowledge for language understanding. 

\subsection{Discussion}

On \control, while our readers outperform our other baselines, they are outperformed by the language modeling baselines from Paperno et al. This suggests that though we have improved the state of the art on LAMBADA by more than 40\% absolute, we have not solved the general language modeling problem; there is no single model that performs well on both \test and \control. Our 36\% estimate of human performance on \control shows the difficulty of the general problem, and reveals a gap of 14\% between the best language model and human accuracy. 

A natural question to ask is whether applying neural readers is a good direction for this task, since they fail on the 17\% of instances which do not have the target word in the context. 
Furthermore, this subset of LAMBADA may in fact display the most interesting and challenging phenomena. 
Some neural readers, like the Stanford Reader, can be easily used to predict target words that do not appear in the context, and the other readers can be modified to do so. Doing this will require a different selection of training data than that used above. 
However, we do wish to note that, in addition to the relative rarity of these instances in LAMBADA, we found them to be challenging for our annotator (who was correct on only 7 of the 13 in this subset). 

We note that \train has similar characteristics to the part of \control that contains the answer in the context (the final column of Table~\ref{accuracy-table}). We find that the ranking of systems according to this column is similar to that in the \test column. This suggests that our simple method of dataset creation could be used to create additional training or evaluation sets for challenging language modeling problems like LAMBADA, perhaps by combining it with baseline suppression~\cite{onishi-16-full}. 

\section{Conclusion}

We constructed a new training set for LAMBADA and used it to train neural readers to improve the state of the art from 7.3\% to 49\%. We also provided results with several other strong baselines 
and included a manual evaluation in an attempt to better understand the phenomena tested by the task. Our hope is that other researchers will seek models and training regimes that simultaneously perform well on both LAMBADA and \control, with the goal of solving the general problem of language modeling. 

\section*{Acknowledgments}
We thank Denis Paperno for answering our questions about the LAMBADA dataset and we thank NVIDIA Corporation for donating GPUs used in this research.

\bibliography{eacl2017}
\bibliographystyle{eacl2017}

\end{document}